\documentclass[10pt,twocolumn,letterpaper]{article}

\usepackage[pagenumbers]{cvpr} % To force page numbers, e.g. for an arXiv version

\usepackage{xspace}  
\newcommand{\datasetname}{\textsc{RPM-10K}\xspace}
\newcommand{\ours}{\textsc{MRLM}\xspace}
\newcommand{\kfm}{\textsc{KFM}\xspace}
\newcommand{\moe}{\textsc{MoE}\xspace}
\newcommand{\refmetric}{\textbf{Ref$\downarrow$}\xspace}
\newcommand{\relmetric}{\textbf{Rel$\downarrow$}\xspace}
\newcommand{\accE}{\textbf{Accuracy$\epsilon$}\xspace}
\newcommand{\accT}{\textbf{Accuracy$\theta$}\xspace}
\newcommand{\benchmarkname}{\textsc{DialBench}\xspace}

\usepackage{graphicx}
\usepackage{times}
\usepackage{helvet}
\usepackage{courier}
\usepackage{amsmath}
\usepackage{algorithm}
\usepackage{makecell}
\usepackage{algorithmic}
\usepackage{enumitem}
\usepackage{csquotes} 
\usepackage{color}
\usepackage{amssymb}
\usepackage{ulem}
\usepackage{float}
\usepackage{multirow}
\usepackage{enumitem}
\usepackage{cite}
\usepackage{mathrsfs}

\usepackage{textcomp,booktabs}
\usepackage{amssymb}% http://ctan.org/pkg/amssymb
\usepackage{pifont}% http://ctan.org/pkg/pifont
\usepackage[misc]{ifsym}

\usepackage{xcolor}
\definecolor{citecolor}{HTML}{0071bc} 
\definecolor{SeaGreen4}{RGB}{0,205,102} 
\definecolor{SlateBlue}{RGB}{106,90,205} 
\definecolor{DarkRed}{RGB}{178,34,34} 
\usepackage[colorlinks, linkcolor=red,  anchorcolor=blue, citecolor=citecolor]{hyperref}
\usepackage{placeins}

\usepackage{colortbl}
\definecolor{mygray}{gray}{.9}
\definecolor{mypink}{rgb}{.99,.91,.95}
\definecolor{mycyan}{cmyk}{.3,0,0,0}

\definecolor{citecolor}{HTML}{0071bc} 
\definecolor{SeaGreen4}{RGB}{0,205,102} 
\definecolor{SlateBlue}{RGB}{106,90,205} 
\definecolor{DarkRed}{RGB}{178,34,34}

\usepackage[capitalize]{cleveref}
\crefname{section}{Sec.}{Secs.}
\Crefname{section}{Section}{Sections}
\Crefname{table}{Table}{Tables}
\crefname{table}{Tab.}{Tabs.}

\usepackage{siunitx}     % for \sisetup and S column type
\usepackage{subcaption}

\usepackage[dvipsnames]{xcolor}

\definecolor{cvprblue}{rgb}{0.21,0.49,0.74}
\def\paperID{12431}      % *** Enter the Paper ID here
\def\confName{CVPR}
\def\confYear{2026}

\title{ DialBench: Towards Accurate Reading Recognition of Pointer Meter using Large Foundation Models }  

\author{Futian Wang$^{1}$, Chaoliu Weng$^{1}$, Xiao Wang$^{1}$\thanks{Corresponding Author: Xiao Wang (xiaowang@ahu.edu.cn)}, Zhen Chen$^{3}$, Zhicheng Zhao$^{2}$, Jin Tang$^{1}$ \\ 
${^1}${School of Computer Science and Technology, Anhui University, Hefei 230601, China} \\ 
${^2}${School of Artificial Intelligence, Anhui University, Hefei 230601, China} \\ 
$^{3}${Department of Computer Science and Information Technology, La Trobe University, Bendigo, Australia} \\ 
\textit{\{wft, xiaowang, zhaozhicheng, tangjin\}@ahu.edu.cn}, \\ \textit{e24201133@stu.ahu.edu.cn, Zhe.Chen@latrobe.edu.au} 
}

\begin{document}
\maketitle

%%%%%%%%% ABSTRACT
\begin{abstract}
\noindent
The precise reading recognition of pointer meters plays a key role in smart power systems, but existing approaches remain fragile due to challenges like reflections, occlusions, dynamic viewing angles, and overly between thin pointers and scale markings. Up to now, this area still lacks large-scale datasets to support the development of robust algorithms. To address these challenges, this paper first presents a new large-scale benchmark dataset for dial reading, termed RPM-10K, which contains 10730 meter images that fully reflect the aforementioned key challenges. Built upon the dataset, we propose a novel vision–language model for pointer meter reading recognition, termed MRLM, based on physical relation injection. Instead of exhaustively learning image-level correlations, MRLM explicitly encodes the geometric and causal relationships between the pointer and the scale, aligning perception with physical reasoning in the spirit of world-model perspectives. Through cross-attentional fusion and adaptive expert selection, the model learns to interpret dial configurations and generate precise numeric readings. Extensive experiments fully validated the effectiveness of our proposed framework on the newly proposed benchmark dataset. Both the dataset and source code will be released on \url{https://github.com/Event-AHU/DialBench}. 
\end{abstract}

\section{Introduction}
%% background of reading recognition of pointer meters 
In industrial and power industries, the reading recognition of pointer meters (such as pressure gauges, ammeters, voltmeters) is critical to ensuring equipment safety and process stability. However, these readings are still predominantly performed through manual, periodic inspections, which suffer from low efficiency, significant subjective errors, frequent omissions or misreadings, inability to enable real-time monitoring, and risks associated with working in hazardous environments. These issues lead to data latency, delayed responses, and a failure to detect transient anomalies, potentially causing equipment overload, energy waste, and even safety accidents, severely hindering the digitalization and intelligent transformation of industrial operations. To address these challenges, it is imperative to develop an AI (Artificial Intelligence)-powered algorithm capable of intelligently and accurately recognizing dial readings. It holds the potential to achieve high-performance automatic recognition, strong generalization capability, end-to-end semantic output, support for multi-modal fusion and contextual understanding, low-cost and rapid deployment, and $7 \times 24h$ continuous monitoring with real-time alerting.

\begin{figure}
\centering
\includegraphics[width=\linewidth]{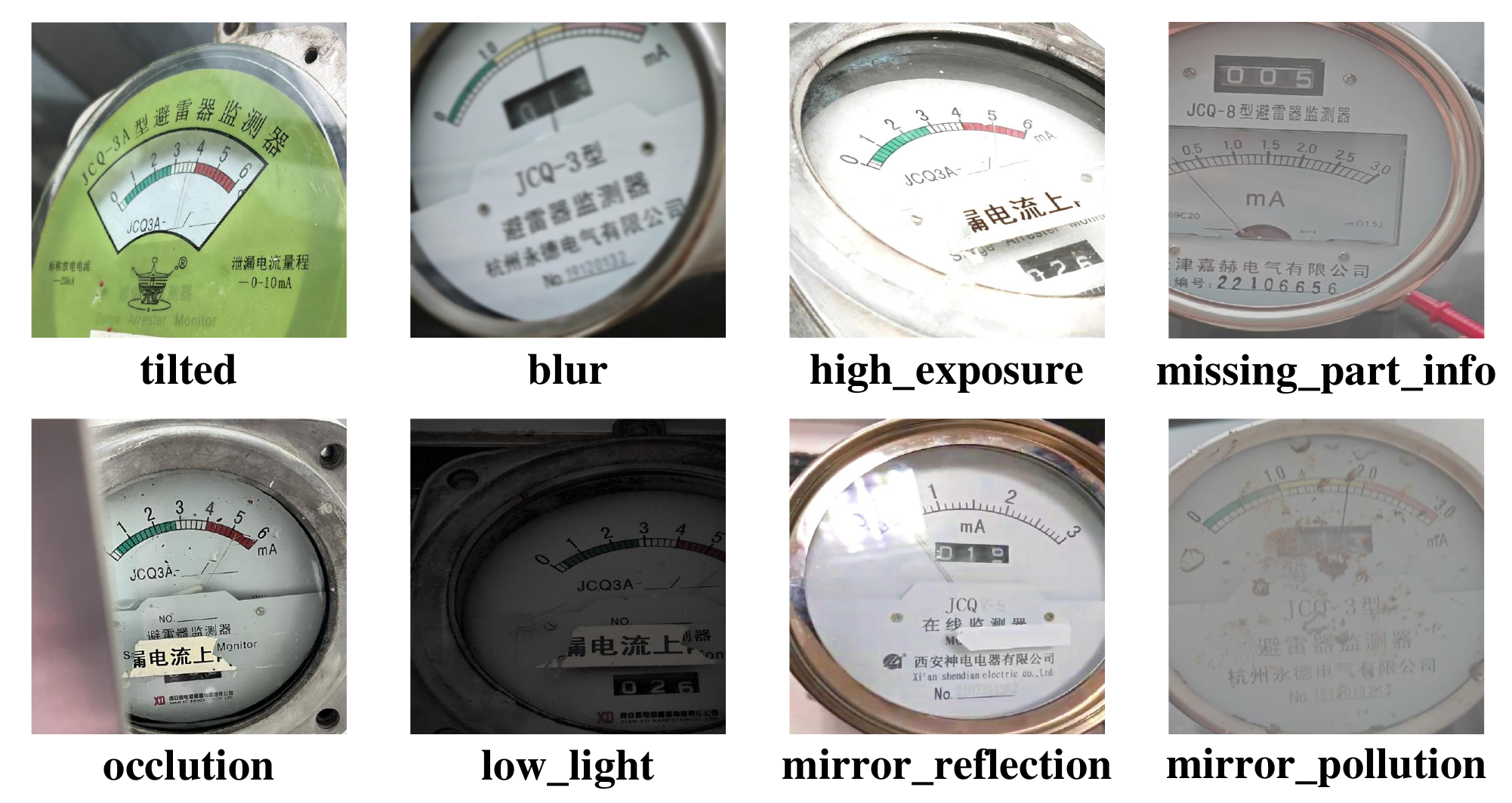}
\caption{Visualization of dials under diverse environmental conditions, including \textit{low\_light}, \textit{high\_exposure}, \textit{missing\_part\_info}, \textit{mirror\_pollution}, \textit{tilted}, \textit{blur}, \textit{occlusion}, and \textit{mirror\_reflection}. These conditions reflect practical challenges for recognition and detection tasks.}
\label{fig:environment_visual}
\end{figure}

Despite the aforementioned features, AI algorithms specifically designed for this task remain extremely scarce~\cite{deepdata}, primarily due to the following reasons:  1) The acquisition of industrial dial images is costly and complex, and often involves enterprise production safety and proprietary information, resulting in an absence of public, large-scale datasets with fine-grained annotations; 2) The diversity of dial types, highly variable industrial environments makes it difficult to develop a unified, generalizable model, hindering scalable deployment and reuse across applications; 3) Critical industries such as power and chemical manufacturing have zero tolerance for misreadings; while traditional rule-based methods offer limited accuracy, they are highly interpretable, whereas the \textit{black-box} nature of AI models often undermines trust and adoption in safety-critical contexts.

%% review of existing works 
Although still scarce, some preliminary efforts have begun to focus on this task. Specifically, Hou et al.~\cite{Hou} proposed a YOLOX-based detection and semantic segmentation framework that accurately localized dial components and achieved a fiducial error below 0.31\%. To enhance robustness under image corruption, Wang et al.~\cite{Wang} introduced a Mask Scoring R-CNN (MSC R-CNN) with image corruption augmentation, a balanced aggregation feature pyramid, and a global context block, attaining a 94\% successful reading rate even in severely degraded conditions. Fan and Li~\cite{Fan} presented an end-to-end approach integrating YOLOv5 for dial detection and an attention-enhanced U2NET for pointer and tick mark extraction, complemented by CRAFT and CRNN for scale recognition, yielding an average reading error below 5\%. More recently, Liu and Shi~\cite{LIU} developed a lightweight YOLOv8S and MC-DeeplabV3Plus-based method that mimics the human reading sequence and incorporates improved attention modules for precise segmentation, achieving fiducial errors of 0.039\% in lab settings and 0.733\% in real-world scenarios. 
Collectively, these studies mark important progress toward intelligent meter reading; however, limitations remain in cross-domain generalization, dataset diversity, and model interpretability, which continue to constrain practical large-scale industrial applications.

%% issues of current works 

%% our work 
To address the aforementioned issues, as shown in Fig.~\ref{fig:environment_visual}, we first propose a large-scale benchmark dataset for the reading recognition of pointer meters, termed RPM-10K, which contains 10730 images and comprehensively captures real-world challenges such as \textit{perspective tilted}, \textit{low light}, \textit{blur}, \textit{occlusion}, \textit{missing part information}, \textit{mirror reflection}, \textit{high exposure}, and \textit{mirror pollution}, across more than 300 dial types, while requiring only text-level annotations to reduce labeling cost. Based on this dataset, we retrain and evaluate 18 multi-modal large models, providing a robust data foundation and standardized benchmark for future algorithmic comparison and analysis. More detailed benchmark results can be found in Table~\ref{tab:main_results}.

Inspired by the tremendous success of large models in natural language processing~\cite{bert, gpt3, llama2} and pre-trained multi-modal models~\cite{vilbert, pali, cambrian}, this paper aims to leverage these foundation models to fully enhance dial reading capability. Given an image of a dial and the physical information (i.e., the scale line), we employ a pre-trained large model CLIP to extract feature representations, and then enhance the current image features through cross-attention. Subsequently, we construct a Mixture-of-Expert (MoE) network capable of handling diverse challenging scenarios such as glare and blur. Meanwhile, the visual features are fed into a Q-Former network to further adapt them into the feature space of a large language model. Finally, a large language model decoder is employed to generate the dial reading as output. An overview of our proposed framework can be found in Fig.~\ref{fig:framework}.

%% contributions 

To sum up, the key contributions of this paper can be summarized as the following three aspects: 

1). We propose a large-scale benchmark dataset for the reading recognition of pointer meter, termed RPM-10K. 18 foundation models are retrained and evaluated on our newly proposed benchmark dataset. The introduction of this benchmark dataset will significantly accelerate the deployment of intelligent dial reading in industrial production. 

2). We propose the first vision–language framework for pointer meter reading that explicitly integrates physical relation injection with a Mixture-of-Experts (MoE) architecture, enabling large models to deeply understand about meter physics. This novel paradigm not only significantly improves reading accuracy and robustness under complex conditions but also establishes a new direction for physics-aware multi-modal AI in industrial vision systems. 

3). Extensive experiments on the newly proposed benchmark dataset fully validate the effectiveness, robustness, and generalizability of our MRLM framework.

% \begin{figure}[!htbp]
%   \centering
%   \includegraphics[width=\columnwidth]{model_general.pdf}
%   \caption{Overview of the model architecture.}
%   \label{fig:model_general}
% \end{figure}

\section{Related Work}
\subsection{Pointer Meter Reading}
With the advancement of smart grids, addressing the challenge of analog meter readings has become a significant research topic. Despite the widespread adoption of digital metering systems, many legacy analog meters remain in use, necessitating efficient and accurate reading methods to facilitate integration into modern smart grid infrastructure. Zuo et al.~\cite{ZUO202090} enhanced the automatic reading of pointer meters by integrating an innovative deep learning algorithm. Specifically, it replaced RoIAlign with PrRoIPooling in the Mask R-CNN framework, improving meter type classification while refining the binary mask fitting for the pointer. The final reading was computed using the proposed angle-based calculation method. Li et al.~\cite{li} employed deep learning techniques to achieve automatic recognition and reading of pointer meters. Initially, they identified and corrected the scale text in the meter image and applied polar transformation to extract the scale region. Subsequently, secondary search and distance measurement were utilized to accurately locate the pointer position, enabling adaptive detection and reading of various pointer meters. Sun et al.~\cite{Sun2023} leveraged deep learning for automatic reading of pointer meters by detecting the meter using YOLOv4 and IFF, segmenting the pointer area with Anam-Net, and recognizing scale text and units using CRAFT and E2E-MLT. Additionally, a lightweight CNN was employed to locate the main scale line in the polar coordinate system. Finally, the reading data was computed based on model outputs, ensuring accurate meter reading. Zhang et al.~\cite{zhang} addressed the challenge of automatic pointer meter reading under motion blur by proposing a one-stage network that integrated deblurring and segmentation. It incorporated high-frequency residual attention to refine detail recovery and employed a judgment-reading algorithm to effectively handle 35 types of meters, ensuring accurate recognition and reading. Xiao et al.~\cite{xiao} improved the automatic reading of circular pointer meters by introducing a robust contour-based perspective rectification scheme. It first estimated a rectification matrix by detecting and fitting the deformed meter contour to suppress noise and then applied the matrix to correct the region of interest, thereby enhancing reading accuracy.

\subsection{Large Foundation Models}
% In the rapidly evolving field of vision-language models (VLMs), several key advancements have focused on instruction-tuned architectures that integrate visual encoders with large language models (LLMs) to enable multimodal reasoning, captioning, and question-answering tasks. 
Early foundational work includes Flamingo~\cite{flamingo}, which pioneered the use of a frozen LLM combined with a perceiver-based visual adapter to process interleaved image-text data, achieving strong few-shot performance on diverse vision-language benchmarks without task-specific fine-tuning.
Building on this, LLaVA~\cite{llava} introduced an instruction-tuned VLM by aligning a vision encoder~\cite{clip} with an LLM like Vicuna through a projection layer, leveraging GPT-4-generated multimodal instruction-following data for end-to-end fine-tuning. This approach emphasized efficient training and demonstrated superior zero-shot generalization on tasks such as visual question answering (VQA) and image captioning. Subsequent models extended these ideas with specialized adaptations. 
InstructBLIP~\cite{instructblip} enhanced the BLIP-2 framework by incorporating instruction-aware fine-tuning, using a Q-Former to bridge vision and language modalities, resulting in improved performance on instruction-following datasets while maintaining computational efficiency.
BLIVA~\cite{bliva} further refined this by integrating BLIP-style bootstrapping with LLaVA-inspired alignment, focusing on better handling of long-context visual inputs and achieving state-of-the-art results in grounded language understanding.
The Qwen-VL series~\cite{qwen1,qwen2,qwen2.5} from Alibaba advanced multilingual and high-resolution capabilities, employing a transformer-based architecture with dynamic visual token compression to support diverse languages and detailed image analysis, making it particularly effective for real-world applications like document understanding.

Datasets and models like ShareGPT-4V~\cite{gpt4v} contributed by providing high-quality, GPT-4V-curated multimodal instruction data, which has been instrumental in fine-tuning VLMs for enhanced visual instruction following, often serving as a backbone for community-driven improvements.
The InternVL series~\cite{internvl1,internvl2,internvl3,internvl4} pushed boundaries with internal scaling laws, combining a large vision foundation model with an LLM for interleaved multimodal processing, achieving competitive results on benchmarks like MM-Vet~\cite{mmvet} and LLaVA-Bench through progressive alignment stages. These models collectively represent a progression from early adapter-based integrations to sophisticated instruction-tuned systems, highlighting trends toward scalability, efficiency, and multimodal coherence in VLMs.

\section{RPM-10K Benchmark Dataset}

\begin{figure*}[!htbp]
\centering
\includegraphics[width=\textwidth]{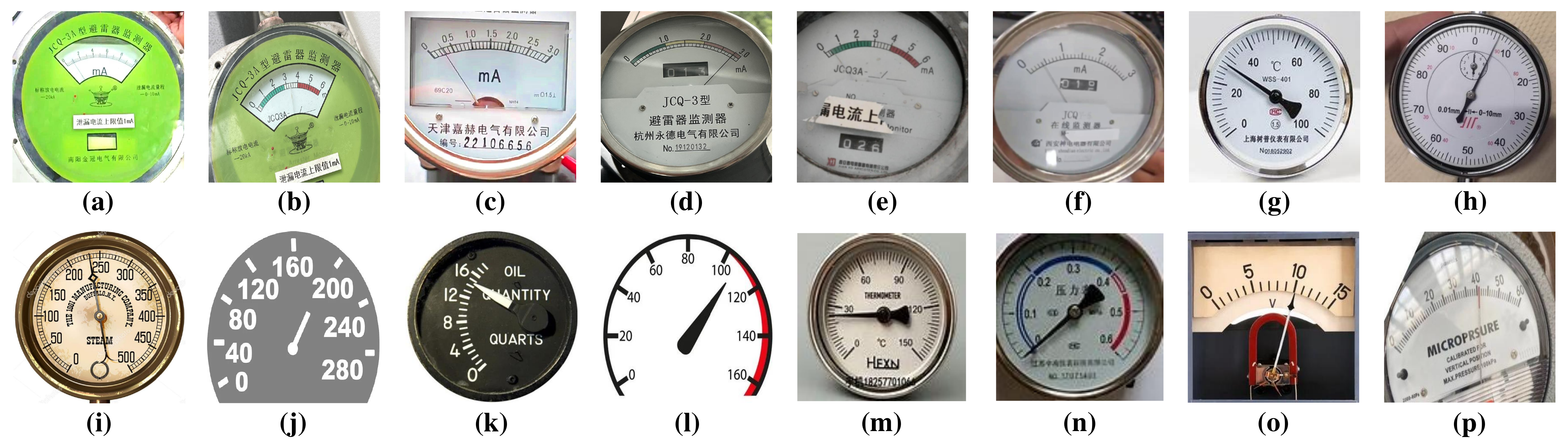}
\caption{Illustration of the dial types in the dataset. Subfigures (a)–(f) represent the six primary dial categories, whereas (g)–(p) correspond to samples acquired from online sources.}
\label{fig:meter_type}
\end{figure*}

To provide an intuitive understanding of the dataset, we present two sets of visualizations. 
Fig.~\ref{fig:meter_type} illustrates representative dial types in the dataset, where subfigures (a)–(f) correspond to the six primary categories of dial types and subfigures (g)–(p) show additional samples collected from web-based sources. Fig.~\ref{fig:environment_visual} highlights the variability of dial appearances under diverse environmental conditions.

\subsection{Protocols} 

To ensure reproducibility and fair comparison, we define standardized protocols for the usage of \datasetname. These protocols guarantee consistent and fair evaluation across different methods and emphasize the novelty of our language-based labeling paradigm for automated meter reading.

\noindent $\bullet$ \textbf{Data Sources:} The dataset consists solely of images paired with language labels, without explicit visual annotations (e.g., bounding boxes or segmentation masks).  

\noindent $\bullet$ \textbf{Diversity of Meter Types:} The dataset covers over 300 types of dials with distinct layouts, scales, and pointer configurations. This diversity significantly increases the difficulty of generalization and requires models to robustly handle heterogeneous dial designs.  

\noindent $\bullet$ \textbf{Environmental Conditions:} Each sample may be affected by one or more of eight complex conditions: low brightness, high exposure, reflection, contamination, blur, tilt, occlusion, and partial information loss. These factors jointly simulate real-world acquisition challenges and stress-test the robustness of dial reading methods.  

\noindent $\bullet$ \textbf{Language Label Annotation Rule:} Instead of conventional visual annotation formats, we directly assign each meter image a textual label representing its reading, e.g., ``\texttt{6.45}''. For multi-dial mechanical instruments, readings from inner/outer scales, mother–child dials, and multi-pointer dials are annotated separately according to their scale or pointer color. Leading zeros and decimal points are preserved to ensure numerical fidelity, enabling end-to-end training of vision–language models without intermediate detection or OCR stages.  

\noindent $\bullet$ \textbf{Train/Test Split:} The dataset is divided into two non-overlapping subsets such that no meter instance appears in more than one split. A typical split ratio of 81\%/19\% is adopted for training and testing, respectively.  

\noindent $\bullet$ \textbf{Evaluation Setting and Usage Protocol:} During evaluation, models receive only raw meter images as input, and their predictions are compared with the ground-truth language labels using the metrics described above. For methods that require fine-tuning, training must be strictly conducted on the designated training split, while the test split is reserved solely for final performance reporting.

\subsection{Data Construction and Annotation} 
The dataset is constructed by extracting frames from meter-reading videos at regular intervals and supplemented with nearly 2,000 web-collected images. All images are manually cropped to retain the dial region and annotated with textual labels. In total, the dataset contains 10,730 meter images, partitioned into 8,730 for training and 2,000 for validation.

\begin{figure}[!htbp]
  \centering
  \includegraphics[width=1\linewidth]{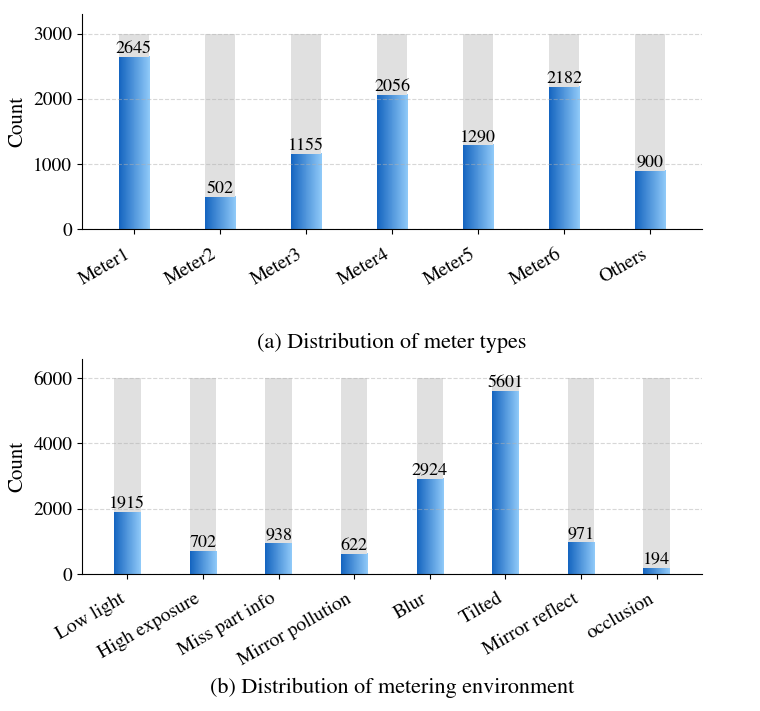}
  \caption{Distribution of samples across different dial configurations and environmental conditions.}
  \label{fig:dial_env_dist}
\end{figure}

\begin{table}[t]
    \centering
    \renewcommand{\arraystretch}{1.2} % <-- 我将这里从 1.0 增加到 1.2 来拉长表格行高
    \begin{tabular}{c c c c}
        \hline
        \textbf{Range} & \textbf{Index Value} & \textbf{\#Dials} & \textbf{\#Samples} \\
        \hline
        10 & 0.2 & 1 & 2{,}645 \\
        3  & 0.1 & 3 & 502 / 1{,}155 / 2{,}056 \\
        6  & 0.2 & 2 & 1{,}290 / 2{,}182 \\
        \hline
    \end{tabular}
    \caption{Dial range and index configurations in the dataset.}
    \label{tab:dial_specs}
\end{table}

\subsection{Statistical Analysis}  
The dataset captures meter readings under controlled variations in dial specifications and environmental conditions, providing a realistic approximation of field scenarios. It covers six major dial types with relatively small measurement ranges, emphasizing fine-grained reading differences.

\noindent $\bullet$ \textbf{Dial Specifications.} 
Most dials feature small measurement ranges.
Table~\ref{tab:dial_specs} summarizes the configurations in terms of range, index value, and sample count.
These settings reflect the focus on compact-range dials frequently encountered in practice, which require precise discrimination between visually similar readings.
Fig.~\ref{fig:dial_env_dist}(a) visualizes the sample distribution across different range–index combinations.

\noindent $\bullet$ \textbf{Visual Distribution Analysis.}  
Fig.~\ref{fig:dial_env_dist}(b) shows the proportion of samples under various environmental conditions.  
\textit{tilted} and \textit{blur} dominate, consistent with their prevalence in real-world deployments.  
Challenging cases such as \textit{missing\_part\_info} and \textit{mirror\_pollution} are less frequent but introduce significant visual degradation.

Overall, the dataset maintains a balanced composition between common and hard cases, facilitating the development of models robust to diverse real-world conditions.

\subsection{Benchmark Baseline}

We evaluate a diverse set of state-of-the-art VLMs as baselines on the proposed \benchmarkname benchmark. 
The evaluated models include Qwen2-VL, InternVL3, LLaMA3.2-Vision, the LLaVA family, BLIVA, and other representative architectures spanning different model sizes and release years. 
These baselines provide a comprehensive reference for assessing the difficulty of \datasetname and the effectiveness of future methods.

\section{Our Proposed Method}

\subsection{Overview} 
\begin{figure}
\centering
\includegraphics[width=\linewidth]{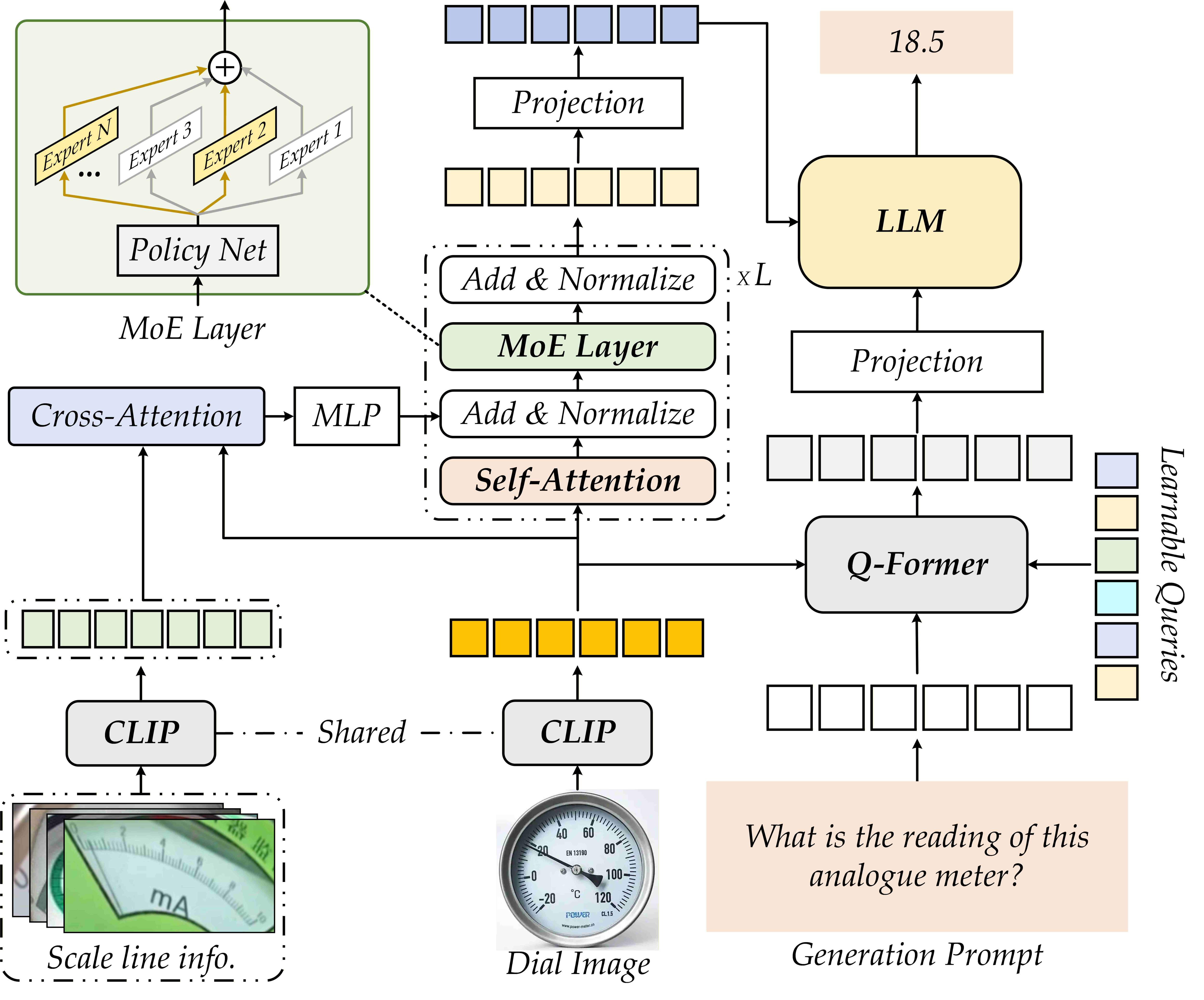}
\caption{Overview of the proposed MeterReading Large Model (MRLM) based on the Physical Relation Injection (PRI) framework. The pipeline sequentially injects physical relations at three hierarchical levels: entity grounding (KFM), relational coupling (MoE), and symbolic alignment (language-labeled supervision).}
\label{fig:framework}
\end{figure}

As shown in Fig.~\ref{fig:framework}, we propose a unified Physical Relation Injection (PRI) framework for high-precision automated meter reading, termed the MeterReading Large Model (MRLM). Unlike conventional visual recognition systems, MRLM explicitly injects physically grounded relations throughout its multi-modal perception pipeline, enabling consistent and physically informed understanding under diverse real-world conditions. The architecture integrates three key innovations hierarchically: the Key Feature Mining (KFM) module grounds perception in physically meaningful entities such as pointers, scales, and digits, enhancing essential components while suppressing background noise; the Mixture-of-Experts (MoE) module models dynamic interactions between meter components, enabling adaptive reasoning under varying visual conditions; and the language-labeled supervision mechanism linguistically constrains the model’s understanding of physical relationships, aligning visual reasoning with symbolic concepts. Together, these components form a comprehensive PRI process that unifies low-level physical grounding, mid-level relational modeling, and high-level symbolic representation to achieve accurate and consistent meter reading.

\subsection{Key Feature Mining}
The \textbf{KFM} module performs the first stage of physical relation injection by grounding the model’s perception on physically meaningful entities. 
Guided by a prior knowledge base of canonical meter archetypes, it isolates and enhances critical dial components, i.e., pointers, scales, and digits, while suppressing irrelevant background.

Template features are first encoded using a pre-trained CLIP model, where early-, mid-, and late-layer features are concatenated:
\begin{equation}
F_i = \text{Concat}(F_{i,\text{early}}, F_{i,\text{mid}}, F_{i,\text{late}}),
\end{equation}
\begin{equation}
F_{\text{template}} = \text{Concat}(F_1, F_2, F_3, F_4, F_5, F_6),
\end{equation}
with $F_{i,\text{early/mid/late}} \in \mathbb{R}^{256 \times 1408}$, yielding $F_{i} \in \mathbb{R}^{256 \times 4224}$ and $F_{\text{template}} \in \mathbb{R}^{1536 \times 4224}$. 
These template features act as the physical query in a cross-attention operation, where the image features $F_{\text{img}} \in \mathbb{R}^{256 \times 4224}$ serve as keys and values:
\begin{equation}
Q = F_{\text{template}}, \quad K = V = F_{\text{img}}.
\end{equation}
The attended features highlight image regions semantically aligned with physical entities:
\begin{equation}
F_{\text{attn}} = \text{softmax}\!\left(\frac{QK^{\top}}{\sqrt{d_k}}\right)V,
\end{equation}
and the final entity-grounded representation is obtained via element-wise fusion:
\begin{equation}
F_{\text{enhanced}} = F_{\text{template}} + F_{\text{attn}}.
\end{equation}
This operation constitutes the first stage of PRI—\textit{injecting entity-level physical grounding} into the model’s visual understanding.

\subsection{Mixture-of-Experts}
The MoE module realizes the second stage of physical relation injection by dynamically modeling physical couplings between meter components.
Given the enhanced representation $x = \text{MLP}(F_{\text{enhanced}}) + F_{\text{img}}$, a gating network $G$ routes the information to $n$ specialized experts $\{E_i\}_{i=1}^{n}$:
\begin{equation}
p = \text{softmax}(G(x)), \quad p \in \mathbb{R}^{n}, \ \sum_{i=1}^{n} p_i = 1.
\end{equation}
Each expert captures distinct coupling relations (e.g., pointer–scale orientation, reflection patterns), and the final fused output is:
\begin{equation}
F_{\text{moe}} = \sum_{i=1}^{n} p_i E_i(x).
\end{equation}
This mechanism enables conditional, physically consistent reasoning under varying conditions—effectively implementing relation-level physical injection across diverse meter types.

\subsection{Language-labeled Supervision}
The final stage of physical relation injection is achieved through language-labeled supervision, which establishes a symbolic bridge between visual relations and linguistic concepts. 
By aligning visual embeddings with structured textual prompts, the model establishes a symbolic correspondence between visual and textual representations.
This symbolic constraint enforces that the numerical predictions generated by the Large Language Model remain consistent with the underlying physical structure of the meter, achieving symbolic-level physical relation injection and enhancing prediction fidelity.

The MRLM is trained end-to-end with a cross-entropy loss:
\begin{equation}
\hat{y} = H(F_{\text{query}}, F_{\text{moe}}, F_{\text{text}}),
\end{equation}
\begin{equation}
\mathcal{L} = - \sum_{c=1}^{C} y_c \log(\hat{y}_c),
\end{equation}
where $H(\cdot)$ denotes the multi-modal fusion network, $F_{\text{query}}$ denotes a set of query vectors from the Q-Former module, $C$ is the number of classes, and $y_c$ is the one-hot ground-truth label for class $c$. This unified training framework ensures that physical relations are injected, preserved, and exploited throughout the full multi-modal inference process.

\section{Experiments}

\begin{table*}[t]
  \centering
  \scriptsize
  \setlength{\tabcolsep}{1pt}
  \renewcommand{\arraystretch}{1.10}
  \renewcommand{\neg}[1]{\textcolor{red}{\bfseries #1}}
  \newcommand{\pos}[1]{\textcolor{teal}{\bfseries #1}}
  \resizebox{0.8\textwidth}{!}{
  \begin{tabular}{l
                  S[table-format=2.1]
                  S[table-format=2.1]
                  c
                  S[table-format=2.1]
                  S[table-format=2.1]
                  c}
    \toprule 
    \multirow{2}{*}{Archetype}
      & \multicolumn{2}{c}{\accE (\%)} & \multirow{2}{*}{\makecell{\(\Delta \accE\)\\(abs, rel)}}
      & \multicolumn{2}{c}{\accT (\%)} & \multirow{2}{*}{\makecell{\(\Delta \accT\)\\(abs, rel)}} \\
    \cmidrule(lr){2-3} \cmidrule(lr){5-6}
      & {with KFM} & {w/o KFM} &  & {with KFM} & {w/o KFM} &  \\
    \midrule
    $meter_1$ & 47.4 & 35.8 & \pos{+11.6 (32.4\%)} & 65.6 & 51.6 & \pos{+14.0 (27.1\%)} \\
    $meter_2$ & 69.6 & 64.8 & \pos{+4.8 (7.4\%)}   & 84.6 & 81.0 & \pos{+3.6 (4.4\%)}  \\
    $meter_3$ & 72.4 & 58.6 & \pos{+13.8 (23.5\%)} & 33.3 & 25.3 & \pos{+8.0 (31.6\%)} \\
    $meter_4$ & 77.9 & 68.8 & \pos{+9.1 (13.2\%)}  & 56.3 & 49.3 & \pos{+7.0 (14.2\%)} \\
    $meter_5$ & 76.1 & 72.6 & \pos{+3.5 (4.8\%)}   & 83.1 & 78.5 & \pos{+4.6 (5.9\%)}  \\
    $meter_6$ & 65.1 & 57.8 & \pos{+7.3 (12.6\%)}  & 75.7 & 68.5 & \pos{+7.2 (10.5\%)} \\
    \midrule
    Average & \textbf{68.1} & \textbf{59.7} &\pos{+8.4 (14.1\%)} & \textbf{66.4} & \textbf{59.0} &\pos{+7.4 (12.5\%)} \\
    \bottomrule
  \end{tabular}
  }
  
  \caption{Performance of different archetypes on \datasetname. The original format “with KFM ± w/o KFM” has been split into separate columns. Absolute and relative improvements are highlighted in green.}
  \label{tab:kfm_archetype_redesign}
\end{table*}
\begin{table*}[t]
  \centering
  \scriptsize
  \setlength{\tabcolsep}{1pt}
  \renewcommand{\arraystretch}{1.10}
  \newcommand{\pos}[1]{\textcolor{teal}{\bfseries #1}}
  \renewcommand{\neg}[1]{\textcolor{red}{\bfseries #1}}
  \resizebox{0.8\textwidth}{!}{
  \begin{tabular}{l
                  S[table-format=2.1]
                  S[table-format=2.1]
                  c
                  S[table-format=2.1]
                  S[table-format=2.1]
                  c}
    \toprule
    \multirow{2}{*}{Environment}
      & \multicolumn{2}{c}{\accE (\%)} & \multirow{2}{*}{\makecell{\(\Delta \accE\)\\(abs, rel)}} 
      & \multicolumn{2}{c}{\accT (\%)} & \multirow{2}{*}{\makecell{\(\Delta \accT\)\\(abs, rel)}} \\
    \cmidrule(lr){2-3} \cmidrule(lr){5-6}
      & {with MoE} & {w/o MoE} & & {with MoE} & {w/o MoE} & \\
    \midrule
    tilted              & 66.5 & 61.6 & \pos{+4.9 (8.0\%)}  & 71.1 & 66.2 & \pos{+4.9 (7.4\%)} \\
    low\_light          & 59.0 & 55.9 & \pos{+3.1 (5.5\%)}  & 66.5 & 65.9 & \pos{+0.6 (0.9\%)} \\
    blur                & 62.8 & 57.1 & \pos{+5.7 (10.0\%)}  & 71.4 & 67.3 & \pos{+4.1 (6.1\%)} \\
    occlusion           & 61.1 & 69.4 & \neg{-8.3 (-12.0\%)} & 61.1 & 72.2 & \neg{-11.1 (-15.4\%)} \\
    missing\_part\_info & 78.3 & 73.7 & \pos{+4.6 (6.2\%)}  & 80.0 & 77.1 & \pos{+2.9 (3.8\%)} \\
    mirror\_reflection  & 64.1 & 60.8 & \pos{+3.3 (5.4\%)}  & 55.8 & 53.0 & \pos{+2.8 (5.3\%)} \\
    high\_exposure      & 58.0 & 48.9 & \pos{+9.1 (18.6\%)} & 67.2 & 66.4 & \pos{+0.8 (+1.2\%)} \\
    mirror\_pollution   & 58.6 & 54.3 & \pos{+4.3 (+7.9\%)} & 64.7 & 67.2  & \neg {-2.5 (-3.7\%)} \\
    \midrule
    Average & \textbf{63.6} & \textbf{60.2} &\pos{+3.4 (+5.6\%)} & \textbf{67.2} & \textbf{66.9} &\pos{+0.3 (+0.4\%)} \\
    \bottomrule
  \end{tabular}
  }
  \caption{MoE performance across environments on \datasetname. The original “with MoE ± w/o MoE” results are split into separate columns. Absolute and relative improvements are shown in green (positive) or red (negative).}
  \label{tab:moe_env_redesign}
\end{table*}

\subsection{Dataset and Evaluation Metric}  
For this study, we utilized \datasetname to train and evaluate our models. The datasets consist of meter readings, where the ground truth values are represented language labels. The Language-based labels allow for end-to-end training, leveraging the model's ability to directly interpret meter readings from visual information.

We use the following evaluation metrics to assess model performance:  

\begin{itemize}
    \item \textbf{Ref}: Calculated as \(\frac{|y - y^*|}{\text{Range}}\), where \(y^*\) is the predicted reading and \(y\) is the true reading. It provides a measure of the absolute error normalized by the range of possible values.
    \item \textbf{Rel}: Calculated as \(\frac{|y - y^*|}{y}\), where \(y^*\) is the predicted reading and \(y\) is the true value. This metric quantifies the relative error of the model’s predictions for each individual meter reading.
    \item \textbf{Accuracy $\epsilon$} (\%): Measures the percentage of predictions where the relative error (ref) is less than or equal to 0.01. This is a crucial metric to assess how close the predicted readings are to the true values within a small error threshold.
    \item \textbf{Accuracy $\theta$} (\%): Measures the percentage of predictions where the angular error (rel) is less than 0.05. This metric evaluates the model’s performance in terms of the angular deviation in readings.
\end{itemize}

These evaluation metrics ensure that the model’s performance is both accurate and robust, providing a comprehensive understanding of its effectiveness in meter reading tasks.

\subsection{Implementation Details}
% Our models adopt a pure language-label paradigm, representing the first end-to-end framework for meter reading trained solely with textual annotations. Unlike conventional pipelines that rely on digit detection, segmentation, or OCR post-processing, our method eliminates these intermediate steps entirely.

% \textbf{Model Architecture.}
% We formulate meter reading as a direct image-to-text generation task: the visual encoder extracts image features, while the language decoder autoregressively produces the reading string. This design removes the need for bounding boxes or handcrafted labels, resulting in a simpler and more scalable architecture.

% \textbf{Training Pipeline.}
% Meter images are paired with precise textual labels, preserving leading zeros and decimals. The model is optimized with cross-entropy loss using AdamW and a cosine learning rate schedule, with mixed-precision training for efficiency. Both the LLM and CLIP backbones are frozen, focusing updates on the connection layers.

% \textbf{Baseline Comparison.}
% For fair comparison, all baselines (Section~\ref{sec:benchmark}) are fine-tuned under identical data splits and optimization settings.

% This framework fully leverages vision-language capabilities while removing the dependence on costly visual annotations, enabling an efficient and scalable solution for automated meter reading.
To balance performance and computational efficiency, all input images are resized to 224 × 224 during both the training and inference stages. A batch size of 8 is used, and the model is optimized with the AdamW optimizer at an initial learning rate of $10^{-4}$. After 200,000 iterations, the learning rate is reduced to $10^{-6}$, followed by an additional 50,000 iterations of fine-tuning to ensure convergence and stability. Our code is implemented using PyTorch~\cite{pytorch} based on Python, and all experiments are conducted on a server equipped with A800 GPUs. More details can be found in our source code.

\begin{table}[t]
\centering
\scriptsize
\setlength{\tabcolsep}{4pt}
\caption{Comparison of pointer meter reading accuracy on the \datasetname dataset. † denotes closed-source visual language model.}
\label{tab:main_results}
\resizebox{\linewidth}{!}{
\begin{tabular}{l l c cccc}
\toprule
No. & Model & Year & Acc $_\epsilon$ (\%) & Acc $_\theta$ (\%) & Ref$\downarrow$ & Rel$\downarrow$ \\
\midrule
01 & Qwen2VL-7B & 2024 & 45.8 & 57.9 & 0.058 & 0.364 \\
02 & Qwen2.5VL-7B & 2025 & 25.6 & 31.3 & 0.31 & 2.28 \\

03 & InternVL3-8B & 2025 & 41.2 & 54.4 & 0.053 & 0.472 \\
04 & LLaMA3.2-Vision-11B~\cite{llama32} & 2024 & 53.6 & 64.6 & 0.046 & 0.287 \\
05 & LLaVA-1.5-7B~\cite{llava1.5} & 2023 & 49.2 & 61.6 & 0.047 & 0.311 \\
06 & LLaVA-v1.6-Vicuna-7B~\cite{llavanext} & 2024 & \underline{56.2} & \underline{67.4} & \textbf{0.038} & \textbf{0.215} \\
07 & MiniCPM-V-2\_6 & 2024 & 45.3 & 57.1 & 0.063 & 0.839 \\
08 & Keye-VL-8B-Preview~\cite{keyevl} & 2025 & 45.5 & 60.4 & 0.043 & 0.239 \\

09 & PaliGemma2-10B-224~\cite{paligemma2} & 2024 & 18.2 & 25.4 & 0.133 & 1.718 \\
10 & Pixtral-12B~\cite{pixtral12b} & 2024 & 36.7 & 50.4 & 0.101 & 1.218 \\
11 & Gemma-3-12B-it~\cite{gemma3} & 2025 & 33.3 & 48.2 & 0.063 & 0.557 \\
12 & Gemma-3n-E4B-Instruct~\cite{gemma3} & 2025 & 7.70 & 9.64 & 0.878 & 6.953 \\
13 & MiniCPM-V-4~\cite{minicpm} & 2024 & 34.3 & 46.6 & 0.073 & 0.607 \\
14 & GLM-4.1V-9B-Thinking~\cite{glm4.1v} & 2025 & 28.8 & 34.6 & 0.172 & 1.282 \\

\midrule
15 & GPT5~\cite{GPT5}$^{\dagger}$ & 2025 & 7.8 & 15.7 & 0.361 & 2.376 \\
16 & Gemini2.5-pro~\cite{Gemini}$^{\dagger}$ & 2025 & 15.8 & 23.4 & 0.213 & 1.088 \\
\midrule
17 & BLIVA~\cite{bliva} & 2023 & 51.2 & 60.0 & 0.114 & 0.925 \\
\rowcolor{mygray}
18 & \textbf{\ours} & 2025 & \textbf{62.4} & \textbf{70.9} & 0.063 & 0.535 \\
\bottomrule
\end{tabular}
}
\end{table}

\subsection{Comparison on Public Benchmark Datasets}
\label{sec:benchmark}
%As shown in Table~\ref{tab:main_results}, our \ours model outperforms all baselines on \datasetname across all metrics. These results demonstrate the effectiveness of integrating KFM and MoE modules into large multimodal models for robust dial reading.
\noindent $\bullet$ \textbf{Analysis of the Accuracy $\epsilon$.}
As shown in Table~\ref{tab:main_results}, under the strict 1\% range normalized error threshold with Ref $\leq 0.01$, \ours achieves the best Accuracy $\epsilon$ at 62.4\%. This surpasses the strongest open source baseline LLaVA v1.6 Vicuna 7B at 56.2\% by 6.2 percentage points, and also exceeds LLaMA3.2 Vision 11B at 53.6\% by 8.8 points and LLaVA 1.5 7B at 49.2\% by 13.2 points. Notably, although some baselines exhibit lower average error, for example LLaVA v1.6 Vicuna 7B reports the lowest Ref at 0.038, their hit rate within the tight 1\% threshold still trails ours, which indicates that \ours concentrates more predictions near the true value and mitigates long tail errors. Closed source models such as GPT5 at 7.8\% and Gemini2.5 pro at 15.8\% lag substantially on this strict metric, which underscores the practical advantage of our approach for high precision meter reading.

\noindent $\bullet$ \textbf{Analysis of the Accuracy $\theta$.}
With the angular deviation tolerance with Rel $< 0.05$, \ours again ranks first, reaching an Accuracy $\theta$ of 70.9\%. This yields consistent gains over LLaVA v1.6 Vicuna 7B at 67.4\% by 3.5 points, LLaMA3.2 Vision 11B at 64.6\% by 6.3 points, and LLaVA 1.5 7B at 61.6\% by 9.3 points, as well as over other recent models such as Keye VL 8B Preview at 60.4\% by 10.5 points and BLIVA at 60.0\% by 10.9 points. While most methods show higher Accuracy $\theta$ than Accuracy $\epsilon$, which reflects the relatively looser angular threshold, \ours maintains the top rank under both criteria, demonstrating robust control of angular errors and strong generalization. Together, the superior Accuracy $\epsilon$ and Accuracy $\theta$ show that \ours reliably delivers strict near exact readings and stable performance within practical angular tolerances.

% ========== 5.4 ==========
\subsection{Component Analysis}
\label{sec:component}
Tables \ref{tab:kfm_archetype_redesign} and \ref{tab:moe_env_redesign} evaluate two modules: \kfm and \moe. Each improves the base pipeline in its target setting, and together they add physical grounding and adaptive routing.
Across six archetypes, \kfm consistently improves \accE and \accT. On average, \accE rises from 59.7\% to 68.1\% and \accT from 59.0\% to 66.4\%. The largest gains appear on $meter_1$ and $meter_3$. Knowledge guided component isolation strengthens entity grounding and suppresses clutter.
Across environments, \moe improves 7 of 8 conditions. Mean \accE increases from 60.2\% to 63.6\% and \accT from 66.9\% to 67.2\%. Excluding occlusion, the gains rise to 5.0 points in \accE and 1.9 points in \accT. The occlusion drop likely reflects limited supervision, and increasing the number of occlusion samples may improve performance.
Overall, \kfm boosts grounding and archetype discrimination, and \moe improves robustness to tilt, blur, and exposure. Combined, they increase accuracy and stability.

\begin{table}[t]
  \centering
  \scriptsize
  \setlength{\tabcolsep}{2pt}
  \resizebox{\columnwidth}{!}{
    \begin{tabular}{l c c c c}
      \toprule
      Variant & {\accE (\%)} & {\accT (\%)} & {\refmetric} & {\relmetric} \\
      \midrule
      \textsc{Baseline}       & 51.2  & 60.0  & 0.114 & 0.925 \\
      w/o \kfm (\moe only)    & 55.4  & 63.4  & 0.079 & \underline{0.469} \\
      w/o \moe (\kfm only)    & \underline{58.3}  \underline& {68.0}  &\textbf{0.056} &\textbf{0.378} \\
      \rowcolor{mygray}
      \textbf{\ours}          &\textbf{62.4}  &\textbf{70.9}  & \underline{0.063} & 0.535 \\
      \bottomrule
    \end{tabular}
  }
  \caption{Ablation study on \datasetname. The table reports performance across four metrics for different variants.}
  \label{tab:ablation1}
\end{table}

\begin{table}[t]
  \centering
  \scriptsize
  \setlength{\tabcolsep}{2pt}
  \resizebox{\columnwidth}{!}{
    \begin{tabular}{l c c c c}
      \toprule
      Variant & {\accE (\%)} & {\accT (\%)} & {\refmetric} & {\relmetric} \\
      \midrule
      \textsc{Baseline}       & 51.2  & 60.0  & 0.114 & 0.925 \\
      4\_scale    & \underline{60.3}  & \underline{68.2}  &\textbf{0.055} & \underline{0.486} \\
      8\_scale    & 58.7  & 65.1  & 0.068 &\textbf{0.368} \\
      6\_expert   & 57.4  & 65.3  & 0.074 & 0.574 \\
      10\_expert   & 59.2  & 67.2  & \underline{0.062} & 0.545 \\
      \rowcolor{mygray}
      \textbf{\ours}         &\textbf{62.4}  &\textbf{70.9}  & 0.063 & 0.535 \\
      \bottomrule
    \end{tabular}
  }
  \caption{This table presents the results of experiments evaluating the parameters of different model modules.}
  \label{tab:ablation2}
\end{table}

% ===================== Model Parameter & Performance Comparison =====================
\begin{table}[t]
  \centering
  \small
  \setlength{\tabcolsep}{1pt}
  \caption{This table presents a comparison of different models in terms of parameter size and inference latency. All measurements are conducted with a batch size of 1 on a single NVIDIA A800 GPU to ensure a fair evaluation of efficiency and computational performance.}
  \label{tab:model-compare}

  \begin{tabular}{
    l
    c
    S[table-format=3.1]
  }
    \toprule
    \multirow{2}{*}{Method} &
    \multicolumn{2}{c}{Efficiency ($\downarrow$)}\\
    \cmidrule(lr){2-3} 
    & {\textbf{Params (M)}} & {\textbf{Latency (ms)}} \\
    \midrule
    \textsc{Baseline} & 7.9B  & 605.6 \\
    w/o \kfm (\moe only)    & 8.6B  & 656.3  \\
    w/o \moe (\kfm only)    & 8.0B  & 580.1  \\
    \rowcolor{mygray}
    \ours &  8.7B & 688.0  \\
    \bottomrule
  \end{tabular}

  \vspace{2pt}
  \raggedright\footnotesize
\end{table}
% ========== 5.5 ==========
\subsection{Ablation Study}
\label{sec:ablation}To assess the contribution of each component, we conduct single-toggle ablations by selectively removing individual modules. Specifically, we consider the following variants: \textbf{w/o \moe}, \textbf{w/o \kfm}, and the full \ours model.
only the target module is toggled. As shown in Table~\ref{tab:ablation1}, both \kfm and \moe contribute positively to overall performance. Removing either component leads to a noticeable performance drop across all evaluation metrics, confirming the complementary roles of the two modules in enhancing model robustness.
We further analyze the parameterization of key modules using the variants in Table~\ref{tab:ablation2}. Here, 4/8scale denote the number of templates in the kfm module, and 6/10expert denote the number of experts in the moe module. Our full model (\ours) uses 6scale and 8expert. Table~\ref{tab:model-compare} shows the parameter counts and latency measured with a batch size of $1$ on a single A800 GPU.
\subsection{Visualization}
\begin{figure}[!htbp]
\centering
\includegraphics[width=\linewidth]{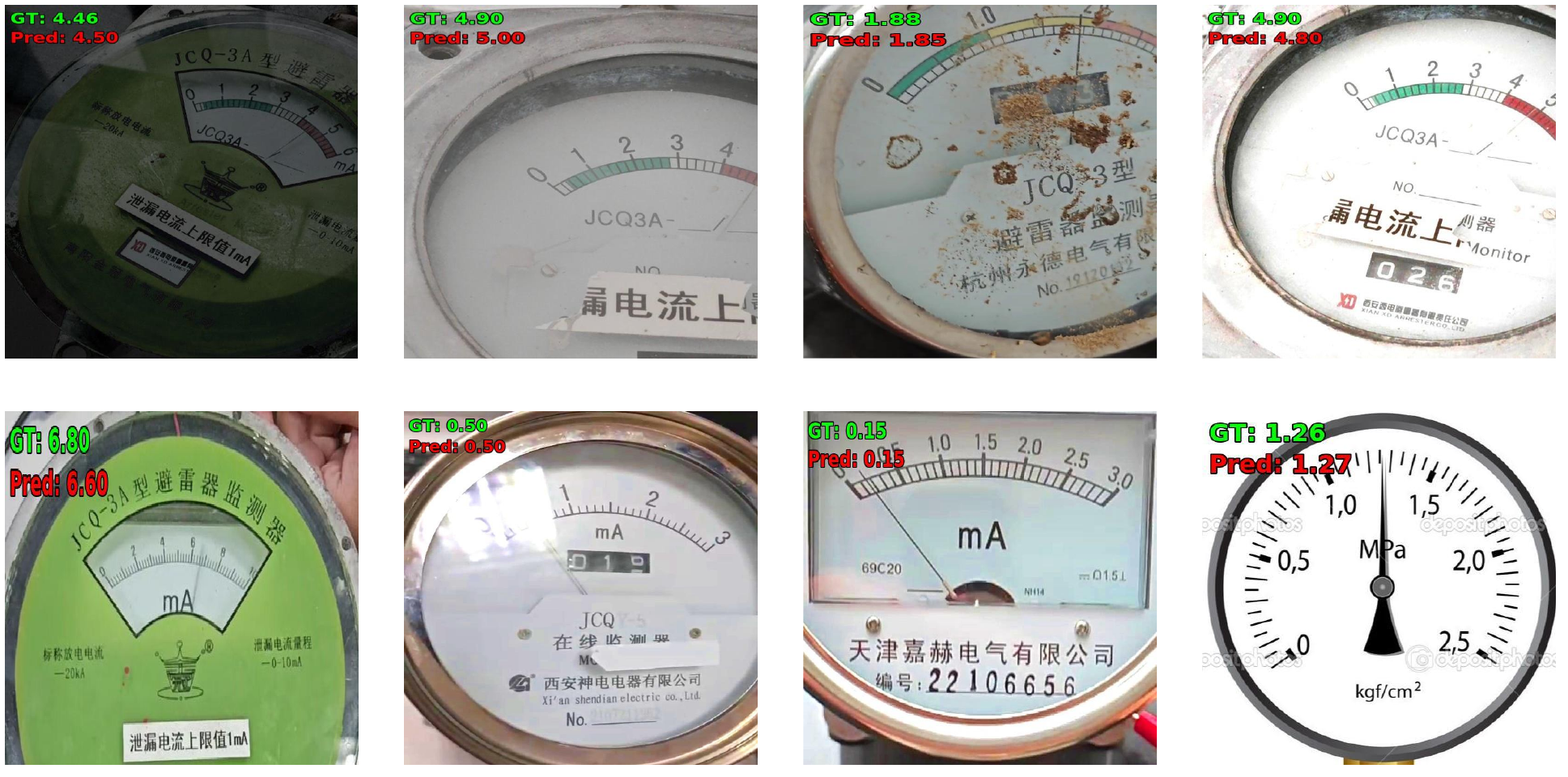}
\caption{Visualization of model prediction results on representative test samples.}
\label{fig:demo}
\end{figure}

To further demonstrate the performance of our method, 
we include a visualization of the model’s prediction results in Fig.~\ref{fig:demo}.
This qualitative analysis complements the quantitative metrics and 
highlights the practical applicability of our approach.
\subsection{Limitation Analysis} 
We have utilized large language models (LLMs), but the expected generalization and emergent capabilities of these models were not observed. The model struggles to predict well on meter images it has never seen before. We initially hypothesized that this limitation stems from the dataset design: the dataset contains only the final meter readings without any explicit reasoning process. To address this, we attempted to augment the dataset with reasoning traces (Chain-of-Thought, CoT) using visual-language models (vLLMs) and prior knowledge. However, the results after supervised fine-tuning (SFT) were even worse. We conjecture that introducing CoT supervision has adverse effects on tasks that are highly sensitive to numerical regression. During training, we observed that the CoT-augmented dataset yielded faster loss convergence and smaller final loss values, which can be attributed to the LLM’s strong capability in predicting the textual part. Nevertheless, its numerical predictions were less accurate, whereas our task is solely concerned with the final numeric reading. We thus hypothesize that the superior performance of the pure numeric dataset arises from the model reducing the meter-reading task to a form of image regression, thereby bypassing the unstable linguistic reasoning process.

\section{Conclusion}
We introduce the MeterReading Large Model, a multimodal framework using Key Feature Mining and a Mixture-of-Experts design for automatic meter reading in complex environments. This approach enhances salient dial features, adapts to diverse meter types, and achieves state-of-the-art performance on our \datasetname benchmark. We also contribute a challenging dataset and standardized benchmark protocols to found future research. Although limitations in generalizing to unseen cases exist, our work provides a robust baseline and opens new directions for applying large multimodal models to industrial perception.

Future work may involve reinforcement learning (RL), which holds great promise for this task. RL has recently shown significant performance boosts for large models in complex reasoning and mathematical problem solving. Incorporating reinforcement learning, may further enhance model accuracy and generalization. This could enable more reliable deployment in diverse real-world scenarios.

{
    \small
    \bibliographystyle{ieeenat_fullname}
    \bibliography{reference}
}

\end{document}